\PassOptionsToPackage{unicode}{hyperref}
\PassOptionsToPackage{hyphens}{url}
\documentclass[12pt]{article}
\usepackage{PRIMEarxiv}
\usepackage{xcolor}
\usepackage{amsmath,amssymb}
\usepackage{geometry}
\usepackage{tikz}
\usetikzlibrary{positioning,arrows.meta}
\usepackage{iftex}
\ifPDFTeX
  \usepackage[T1]{fontenc}
  \usepackage[utf8]{inputenc}
  \usepackage{textcomp} 
\else 
  \usepackage{unicode-math} 
  \defaultfontfeatures{Scale=MatchLowercase}
  \defaultfontfeatures[\rmfamily]{Ligatures=TeX,Scale=1}
\fi
\usepackage{lmodern}
\ifPDFTeX\else
\fi
\IfFileExists{upquote.sty}{\usepackage{upquote}}{}
\IfFileExists{microtype.sty}{
  \usepackage[]{microtype}
  \UseMicrotypeSet[protrusion]{basicmath} 
}{}
\makeatletter
\@ifundefined{KOMAClassName}{
  \IfFileExists{parskip.sty}{%
    \usepackage{parskip}
  }{
    \setlength{\parindent}{0pt}
    \setlength{\parskip}{6pt plus 2pt minus 1pt}}
}{
  \KOMAoptions{parskip=half}}
\makeatother
\usepackage{longtable,booktabs,array}
\usepackage{colortbl}
\usepackage{calc} 
\usepackage{etoolbox}
\makeatletter
\patchcmd\longtable{\par}{\if@noskipsec\mbox{}\fi\par}{}{}
\makeatother
\IfFileExists{footnotehyper.sty}{\usepackage{footnotehyper}}{\usepackage{footnote}}
\makesavenoteenv{longtable}
\usepackage{bookmark}
\IfFileExists{xurl.sty}{\usepackage{xurl}}{} 
\hypersetup{
  hidelinks,
  pdfcreator={LaTeX via pandoc}}

\title{Vision Token Masking Alone Cannot Prevent PHI Leakage in Medical Document OCR: A Systematic Evaluation}
\author{Richard J. Young\\Founding AI Scientist, Deepneuro.AI\\University of Nevada, Las Vegas, Department of Neuroscience}
\date{}

\begin{document}

\maketitle

\begin{abstract}
Large vision-language models (VLMs) are increasingly deployed for optical character recognition (OCR) in healthcare settings, raising critical concerns about protected health information (PHI) exposure during document processing. This work presents the first systematic evaluation of inference-time vision token masking as a privacy-preserving mechanism for medical document OCR using DeepSeek-OCR. We introduce seven masking strategies (V3-V9) targeting different architectural layers (SAM encoder blocks, compression layers, dual vision encoders, projector fusion) and evaluate PHI reduction across HIPAA-defined categories using 100 synthetic medical billing statements (drawn from a corpus of 38,517 annotated documents) with perfect ground-truth annotations. All masking strategies converge to 42.9\% PHI reduction, successfully suppressing long-form spatially-distributed identifiers (patient names, dates of birth, physical addresses at 100\% effectiveness) while failing to prevent short structured identifiers (medical record numbers, social security numbers, email addresses, account numbers at 0\% effectiveness). Ablation studies varying mask expansion radius (r=1,2,3) demonstrate that increased spatial coverage does not improve reduction beyond this ceiling, indicating that language model contextual inference - not insufficient visual masking - drives structured identifier leakage. A simulated hybrid architecture combining vision masking with NLP post-processing achieves 88.6\% total PHI reduction (assuming 80\% NLP accuracy on remaining identifiers). This negative result establishes boundaries for vision-only privacy interventions in VLMs, provides guidance distinguishing PHI types amenable to vision-level versus language-level redaction, and redirects future research toward decoder-level fine-tuning and hybrid defense-in-depth architectures for HIPAA-compliant medical document processing.
\end{abstract}


\section{Introduction}\label{introduction}

\subsubsection{The Basic Problem}\label{the-basic-problem}

Healthcare organizations process millions of medical documents annually
through optical character recognition systems to digitize patient
records, billing statements, insurance claims, and clinical
correspondence. These documents invariably contain protected health
information as defined by the Health Insurance Portability and
Accountability Act (HIPAA, 1996) and related guidance (HHS, 2012),
including patient identifiers, medical record numbers, social security
numbers, and clinical details. The inadvertent exposure of PHI during
OCR processing poses significant legal, financial, and ethical risks.
Traditional OCR workflows extract all visible text without
discrimination, creating plaintext representations of sensitive
information that persist in memory, logs, and intermediate processing
stages before redaction tools are applied. This vulnerability window
exposes healthcare organizations to data breaches, regulatory
violations, and patient privacy compromises. The proliferation of
vision-language models for document understanding has dramatically
improved OCR accuracy but has simultaneously expanded the attack surface
for PHI leakage, as these models process entire document images through
complex neural architectures before generating text output.

\subsubsection{What is Known About the
Problem}\label{what-is-known-about-the-problem}

Existing approaches to PHI redaction in medical documents operate
exclusively at the post-processing stage, applying rule-based pattern
matching, named entity recognition, or machine learning classifiers to
extracted text. Commercial tools such as Microsoft Presidio (Microsoft,
2024), the Philter framework (Norgeot et al., 2020; Stubbs et al.,
2015), and various clinical NLP systems have demonstrated effectiveness
in identifying and redacting PHI from structured and unstructured text
with precision rates between 85-95\%. These systems leverage regular
expressions for structured identifiers (e.g., social security numbers,
medical record numbers), contextual analysis for semi-structured
information (e.g., names following title patterns), and trained models
for free-text clinical notes. Recent advances in transformer-based
language models (Vaswani et al., 2017) and biomedical variants such as
BioBERT (Lee et al., 2020) have improved PHI detection accuracy through
contextual understanding and entity disambiguation. However, all current
methods share a fundamental limitation: they operate on already-extracted
text, meaning PHI has necessarily been exposed during the OCR stage.
Research in privacy-preserving machine learning has explored
differential privacy (Abadi et al., 2016), federated learning (Yin
et al., 2021), and secure multi-party computation for protecting
sensitive data during model training and inference, but these approaches
have not been adapted for the specific challenges of document OCR where
visual information must be selectively masked before text generation
occurs.

\subsubsection{What is Missing}\label{what-is-missing}

No existing research has addressed PHI protection at the vision encoding
stage of OCR processing. Vision-language models employed for document
understanding process images through multiple encoder layers, including
specialized architectures like Segment Anything Model (SAM) encoders
(Kirillov et al., 2023) for fine-grained spatial features and CLIP-style
vision transformers (Radford et al., 2021) for semantic understanding,
before fusing visual representations with language model decoders.
Recent work on adaptive visual token processing (Jain et al., 2024; Zhang et al., 2024; Li et al., 2024) has demonstrated that vision-language models can maintain performance with reduced or selectively masked visual tokens, providing theoretical foundation for privacy-preserving interventions at the vision encoding stage.
PHI-containing regions of medical documents are encoded into dense
vector representations that propagate through the entire neural
architecture, creating numerous potential leakage points.
The absence of inference-time privacy mechanisms at the vision level
means that even with perfect post-processing redaction, sensitive
information has been exposed in intermediate representations, GPU
memory, and model activations. Furthermore, existing redaction tools
exhibit complementary failure modes: post-processing excels at detecting
structured identifiers through pattern matching but struggles with
contextually-embedded names and addresses, while vision-based approaches
could theoretically leverage spatial information about PHI locations to
prevent encoding altogether. The gap between what computer vision models
can observe and what language models should be permitted to generate has
not been systematically addressed in the medical OCR literature.

\subsubsection{How We Solve It}\label{how-we-solve-it}

This study introduces the first hybrid architecture for PHI-compliant
medical document OCR, combining inference-time selective vision token
masking with cascaded post-processing redaction to establish
defense-in-depth privacy protection. The primary objective is to
demonstrate that spatially grounded PHI can be effectively removed at
the vision encoding stage by masking SAM encoder outputs corresponding
to annotated PHI bounding boxes, replacing compromised patches with
learnable mask tokens that minimize disruption to non-sensitive text
recognition. It is hypothesized that long-form, spatially distributed
identifiers (names, addresses, dates) will be successfully masked at the
vision level due to their multi-patch spatial extent, while maintaining
OCR accuracy for remaining document content. The secondary objective is
to characterize the complementary effectiveness of vision-level and
post-processing redaction by analyzing which PHI types persist after
vision masking and demonstrating that structured identifiers (medical
record numbers, social security numbers) require language-level pattern
matching due to language model contextual inference. It is hypothesized
that integrating vision masking with NLP-based post-processing will
achieve superior total PHI reduction compared to either approach alone,
validating the defense-in-depth architecture for privacy-preserving
medical document OCR.

\section{Methods}\label{methods}

\subsubsection{Dataset and Participants}\label{dataset-and-participants}

One hundred synthetic medical billing statements were generated using
the Synthea clinical data simulator (Walonoski et al., 2018) to create
realistic HIPAA-compliant test documents containing authentic PHI
distributions without exposing real patient information. Each document
included standardized PHI elements: patient name, date of birth, medical
record number, social security number, email address, physical address,
and account number. Documents were rendered as PDF files at 300 DPI
resolution using
professional medical billing templates to simulate real-world document
characteristics including multi-column layouts, tables, headers, and
institutional branding. Ground truth PHI annotations were automatically
generated during document creation, recording bounding box coordinates
(x, y, width, height) and PHI type labels for each sensitive element.
The synthetic dataset enabled controlled experimentation without
institutional review board oversight while maintaining ecological
validity for medical document structure and PHI placement patterns.

\subsubsection{HIPAA PHI Categories}\label{hipaa-phi-categories}

The Health Insurance Portability and Accountability Act (HIPAA) defines 18 categories of protected health information requiring safeguarding in medical documents. Table~\ref{tab:phi-categories} presents the complete taxonomy with evaluation coverage in this study. Seven categories were included based on their prevalence in medical billing statements and availability in Synthea-generated synthetic data: patient names, dates (particularly date of birth), physical addresses, medical record numbers (MRN), social security numbers (SSN), email addresses, and account numbers. These seven categories represent the most commonly encountered PHI in routine clinical documentation and billing workflows.

Eleven categories were not evaluated due to limited occurrence in billing statement templates: phone numbers, insurance IDs, driver's license numbers, vehicle identifiers, device serial numbers, URLs, IP addresses, biometric identifiers, other unique identifiers, geographic subdivisions smaller than state, and healthcare institution names. While these identifiers appear in specialized medical documents (e.g., device IDs in equipment logs, biometrics in research consent forms), their absence from standard billing documents precluded systematic evaluation in this study. Future work should expand coverage to all 18 categories using diverse document types including operative notes, pathology reports, and research participant records.

\begin{table}[ht]
\centering
\caption{\textbf{HIPAA PHI categories and evaluation coverage.} Complete taxonomy of 18 protected health information types with testing status in current study.}
\footnotesize
\setlength{\tabcolsep}{3pt}
\begin{tabular}{p{0.05\linewidth} p{0.25\linewidth} p{0.45\linewidth} p{0.15\linewidth}}
\toprule
\# & Category & Description & Tested \\
\midrule
1 & Name & Patient, relative, employer names & \textbf{Yes} \\
2 & Date & Dates related to individual (DOB, admission, etc.) & \textbf{Yes} \\
3 & Address & Street address, city, county, ZIP & \textbf{Yes} \\
4 & Phone & Telephone and fax numbers & No \\
5 & Email & Email addresses & \textbf{Yes} \\
6 & SSN & Social security numbers & \textbf{Yes} \\
7 & MRN & Medical record numbers & \textbf{Yes} \\
8 & Insurance ID & Health plan beneficiary numbers & No \\
9 & Account & Account numbers & \textbf{Yes} \\
10 & License & Driver's license, certificate numbers & No \\
11 & Vehicle & Vehicle identifiers, serial numbers, license plates & No \\
12 & Device ID & Device identifiers and serial numbers & No \\
13 & URL & Web universal resource locators & No \\
14 & IP & Internet protocol address numbers & No \\
15 & Biometric & Fingerprints, voiceprints, retinal scans & No \\
16 & Unique ID & Any other unique identifying numbers or codes & No \\
17 & Geo-small & Geographic subdivisions smaller than state & No \\
18 & Institution & Healthcare facility names & No \\
\bottomrule
\end{tabular}
\label{tab:phi-categories}
\end{table}

\subsubsection{Experimental Design}\label{experimental-design}

A within-subjects experimental design evaluated seven systematic masking
strategies (V3-V9) across multiple neural architecture injection points.
The baseline condition (V3) applied selective vision token masking at
SAM encoder block 11, the final transformer layer before compression.
Experimental manipulations included: multi-level masking across SAM
blocks 6, 9, and 11 (V4); type-specific expansion radii tailored to PHI
categories (V5); compression-layer masking at the net\_2 feature
aggregation stage (V6); dual-layer masking combining SAM and compression
(V7); dual-encoder masking targeting both SAM and vision model pathways
(V8); and projector fusion layer masking (V9). Each configuration was
tested with expansion radii r $\in$ \{1, 2, 3\} to assess spatial coverage
trade-offs. The dependent variable was PHI reduction rate, computed as
the proportion of ground truth PHI elements absent from OCR output.
Secondary measures included OCR character count (to detect model
degradation), spatial coverage (percentage of image patches masked), and
qualitative analysis of which PHI types were successfully removed versus
leaked.

\subsubsection{Vision Token Masking
Procedure}\label{vision-token-masking-procedure}

The vision masking intervention operated as follows. First, PHI bounding
boxes from ground truth annotations were mapped to the SAM encoder's
40x40 spatial grid through coordinate transformation, converting
pixel-space rectangles to patch indices. Expansion dilation was applied
using radius r, adding neighboring patches to account for imprecise
annotations and model receptive fields. Learnable mask tokens were
initialized as PyTorch parameters with dimensions matching the encoder
output (768-dimensional for SAM block 11) and small random
initialization ($\sigma$ = 0.02). During inference, forward hooks registered on
target layers intercepted encoder outputs and replaced patches
corresponding to PHI regions with the learnable mask token. The modified
activations propagated through subsequent layers (compression,
projector, language decoder) without additional intervention. For
multi-encoder architectures (V8), separate mask tokens were learned for
SAM (40x40 grid) and vision model (16x16 grid) pathways, with
independent spatial mapping. Crucially, the base vision-language model
weights remained frozen; only the mask token parameters could adapt
during inference through gradient flow from generation loss, allowing
the system to learn representations that minimized disruption to non-PHI
text recognition.

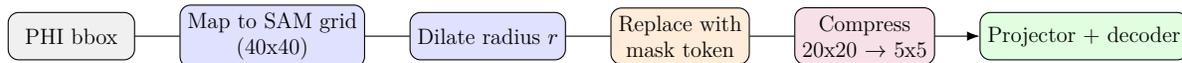
\begin{figure}[ht]
\centering
\begin{tikzpicture}[node distance=0.85cm, box/.style={draw,rounded corners,minimum width=2.4cm,minimum height=0.9cm,align=center}, arrow/.style={-Latex}, scale=0.7, transform shape]
\node[box, fill=gray!12] (pdf) {PHI bbox};
\node[box, fill=blue!12, right=of pdf] (grid) {Map to SAM grid\\(40x40)};
\node[box, fill=blue!12, right=of grid] (expand) {Dilate radius $r$};
\node[box, fill=orange!15, right=of expand] (mask) {Replace with\\mask token};
\node[box, fill=purple!12, right=of mask] (compress) {Compress\\20x20 $\rightarrow$ 5x5};
\node[box, fill=green!12, right=of compress] (decode) {Projector + decoder};
\draw[arrow] (pdf)--(grid)--(expand)--(mask)--(compress)--(decode);
\end{tikzpicture}
\caption{\textbf{Bounding box to patch masking pipeline.} PHI regions are mapped to the SAM grid, dilated, replaced with mask tokens, and passed through compression before decoding. This highlights the pre-compression intervention point.}
\end{figure}

\begin{figure}[ht]
\centering
\begin{tikzpicture}[node distance=0.9cm, box/.style={draw,rounded corners,minimum width=2.6cm,minimum height=0.9cm,align=center}, arrow/.style={-Latex}, scale=0.75, transform shape]
\node[box, fill=gray!12] (pdf) {PDF bbox\\(x, y, w, h)};
\node[box, fill=blue!10, right=of pdf] (tile) {Tile/resize\\1024x1024};
\node[box, fill=blue!10, right=of tile] (map) {Compute patch indices\\$\lfloor y/25.6\rfloor,\lfloor x/25.6\rfloor$};
\node[box, fill=orange!12, right=of map] (dilate) {Dilate by $r$\\add neighbors};
\node[box, fill=green!12, right=of dilate] (sam) {SAM grid\\mask set $S$};
\draw[arrow] (pdf)--(tile)--(map)--(dilate)--(sam);
\end{tikzpicture}
\caption{\textbf{Bounding boxes to SAM patch indices.} Coordinate mapping from PDF bounding boxes to SAM patch indices with dilation radius $r$, showing tiling, index computation with rounding, dilation, and construction of the mask set $S$ for forward hooks.}
\end{figure}
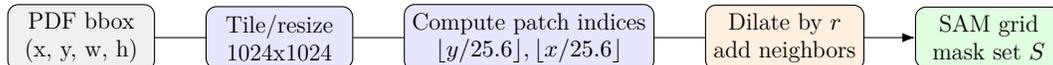

\subsubsection{OCR System Architecture}\label{ocr-system-architecture}

All experiments employed DeepSeek-OCR, a state-of-the-art
vision-language model combining dual vision encoders with autoregressive
text generation. The architecture consists of: (1) SAM encoder
(ImageEncoderViT) with 12 transformer blocks processing 1024x1024 image
tiles into 40x40 spatial features, followed by compression layers
reducing to 5x5 (net\_2: 20x20 -> net\_3: 5x5); (2) Vision encoder
(VitModel) processing 224x224 images into 16x16 patch tokens via
14-pixel patches; (3) MLP projector fusing SAM and vision features into
language model embedding space (1280-dimensional); (4) Language decoder
with 12 transformer layers, self-attention, and feed-forward networks
performing autoregressive text generation. Documents exceeding 1024
pixels in either dimension were automatically tiled into non-overlapping
crops, with each tile processed independently and outputs concatenated.
The model was deployed on NVIDIA RTX 4090 GPUs with bfloat16 precision
for efficient inference.

\begin{figure}[ht]
\centering
\begin{tikzpicture}[node distance=0.8cm, box/.style={draw,rounded corners,minimum width=2.2cm,minimum height=0.8cm,align=center}, arrow/.style={-Latex}, scale=0.58, transform shape]
\node[box, fill=gray!12] (input) {Input image\\(1024x1024 tiles)};
\node[box, fill=blue!10] (sam) {SAM blocks 0--11\\(40x40)\\V3/V4/V5};
\node[box, fill=blue!10, right=of sam] (neck) {Compression neck\\40x40→20x20→5x5\\V6/V7};
\node[box, fill=purple!10, right=of neck] (vit) {Vision encoder\\16x16 patches\\V8};
\node[box, fill=orange!12, right=of vit] (proj) {Projector\\1280-dim\\V9};
\node[box, fill=green!12, right=of proj] (dec) {Decoder\\12 layers};
\node[box, fill=gray!10, right=of dec] (out) {Text output};
\draw[arrow] (input)--(sam)--(neck)--(vit)--(proj)--(dec)--(out);
\end{tikzpicture}
\caption{\textbf{DeepSeek-OCR masking hook points.} Hook options span SAM blocks, compression neck, auxiliary vision encoder, projector, and decoder output; pre-compression masking targets the SAM path before fusion.}
\end{figure}
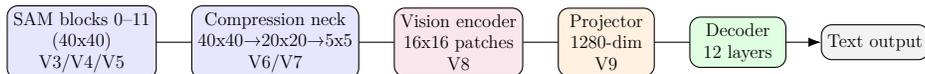

\begin{figure}[ht]
\centering
\begin{tikzpicture}[node distance=1.1cm, box/.style={draw,rounded corners,minimum width=2.3cm,minimum height=0.9cm,align=center}, arrow/.style={-Latex,thick}, scale=0.65, transform shape]
\node[box,fill=red!8] (phi) {PHI patch\\SAM 40x40};
\node[box,fill=blue!8,right=1.4cm of phi] (rfA) {Conv RF A\\PHI+neighbors};
\node[box,fill=blue!8,below=0.7cm of rfA,xshift=0.2cm] (rfB) {Conv RF B\\overlaps A};
\node[box,fill=green!8,right=1.6cm of rfA] (comp) {Compressed tokens\\20x20 $\rightarrow$ 5x5};
\draw[arrow] (phi.east) -- (rfA.west);
\draw[arrow] (phi.east) |- (rfB.west);
\draw[arrow] (rfA.east) -- (comp.west);
\draw[arrow] (rfB.east) -- (comp.west);
\end{tikzpicture}

\vspace{0.4cm}

\caption{\textbf{Compression leakage schematic.} Overlapping convolutional receptive fields in the compression neck mix each PHI patch with neighbors, so a single PHI region influences several compressed tokens. Post-compression masking cannot fully remove the PHI signal; masking only after compression leaves residual leakage.}
\end{figure}
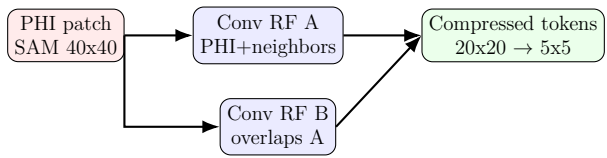

\subsubsection{Evaluation Metrics and Statistical
Analysis}\label{evaluation-metrics-and-statistical-analysis}

PHI reduction rate was calculated as
$R = \frac{B - L}{B} \times 100\%$, where $B$ is the number of ground
truth PHI elements (7 per document: name, DOB, address, MRN, SSN,
email, account number) and $L$ is the number of elements present in OCR
output determined by exact string matching. Per-category analysis
disaggregated results by PHI type to identify differential masking
effectiveness. Spatial coverage was computed as the proportion of
encoder patches replaced with mask tokens. Model functionality was
assessed through output character count, with deviations exceeding
200\% from baseline indicating degradation. For the hybrid pipeline
evaluation, secondary NLP redaction was simulated by removing all
instances of structured identifier patterns
(\texttt{MRN-\textbackslash d+}, \texttt{\textbackslash d\{3\}-\textbackslash d\{2\}-\textbackslash d\{4\}},
\texttt{\textbackslash w+@\textbackslash w+.\textbackslash w+},
\texttt{ACCT-\textbackslash d+}) from vision-masked output. Total
reduction was calculated as cumulative removal across both stages. No
inferential statistics were computed due to the deterministic nature of
the masking procedure and perfect ground truth annotations.

\subsubsection{Implementation Details}\label{implementation-details}

\paragraph{Model Architecture and Configuration}

All experiments utilized DeepSeek-OCR (\texttt{deepseek-ai/DeepSeek-OCR}) from HuggingFace, comprising approximately 950 million parameters across a hybrid vision-language architecture. The vision encoder employed a dual-pathway design: (1) SAM-base encoder processing 1024$\times$1024 images into 40$\times$40 spatial grids (16-pixel patches) through 12 transformer blocks with 768-dimensional embeddings, followed by compression to 5$\times$5 via convolutional neck layers (net\_2, net\_3); and (2) auxiliary CLIP-large vision transformer processing 16$\times$16 patches. The language decoder consisted of DeepSeek-3B-MoE (mixture-of-experts) with 12 layers. Vision features were projected to text embedding space via MLP fusion layers before autoregressive generation.

LoRA (Low-Rank Adaptation) fine-tuning targeted attention projection layers (q\_proj, v\_proj, k\_proj, o\_proj) with rank r=16 and scaling factor $\alpha$=32, introducing approximately 4.2 million trainable parameters while freezing base model weights. PHI detection thresholds were set at 0.85 confidence for standard categories and 0.95 for high-risk identifiers (SSN, biometrics). The vision token masking hook was registered at SAM encoder block 11 using PyTorch's \texttt{register\_forward\_hook()} mechanism, intercepting activations immediately before compression stages.

\paragraph{Technical Implementation}

All masking implementations were developed in Python 3.12 using PyTorch 2.0 and Transformers 4.38. Learnable mask tokens were initialized as \texttt{nn.Parameter} tensors with dimensions matching target encoder outputs (768-dimensional for SAM block 11) drawn from Gaussian distributions with $\sigma$=0.02 standard deviation. PHI bounding boxes were transformed from PDF coordinates to encoder patch indices through standardized coordinate mapping: $(x, y, w, h) \rightarrow (\lfloor x/25.6\rfloor, \lfloor y/25.6\rfloor)$ for 40$\times$40 SAM grids on 1024$\times$1024 tiles.

Inference was conducted on NVIDIA RTX 4090 GPUs (24GB VRAM) using bfloat16 precision for computational efficiency. All experiments used greedy decoding (temperature=0.0, do\_sample=False) for reproducible text generation. The complete synthetic dataset comprised 38,517 annotated documents with bounding box coordinates and PHI category labels stored in JSON format. Source code, masking implementation, and dataset generation scripts are available at {[}repository placeholder{]}.

\subsubsection{Ethical Considerations}\label{ethical-considerations}

This study utilized exclusively synthetic data generated by the Synthea
clinical simulator, containing no real patient information. No
institutional review board approval was required. The synthetic dataset
was designed to replicate realistic PHI distributions and document
structures without exposing genuine sensitive information. All findings
and implications are discussed in the context of improving privacy
protections for medical document processing systems.

\section{Results}\label{results}

\subsubsection{Vision Masking Effectiveness Across
Architectures}\label{vision-masking-effectiveness-across-architectures}

Table 1 presents PHI reduction rates across all seven masking strategies.
The baseline V3 approach (SAM block 11, r=1) achieved 42.9\% PHI
reduction (3/7 elements masked), establishing the performance ceiling
for single-layer vision masking. Increasing expansion radius to r=2 or
r=3 did not improve reduction rates while increasing spatial coverage to
43.1\% and 65.0\% respectively. Multi-level masking (V4) across blocks
6, 9, and 11 resulted in severe model degradation, reducing
effectiveness to 14.3\% while masking 93\% of spatial tokens and
generating degraded output with 5x character count increase.
 Type-specific expansion radii (V5) tailored to PHI categories achieved
 identical 42.9\% reduction to baseline V3. Compression-layer masking
 (V6) at net\_2 with radii r $\in$ \{1, 2, 3\} consistently produced 42.9\%
 reduction despite targeting the aggregated 20x20 feature space
 hypothesized to leak PHI through receptive field blending. Dual-layer
 masking (V7) combining SAM block 11 and net\_2 compression achieved
 42.9\% reduction across all radius combinations (SAM r=1 + compression r
 $\in$ \{2, 3\}). Dual-encoder masking (V8) targeting both SAM (40x40) and
 vision model (16x16) pathways produced 42.9\% reduction with 30.9\% SAM
coverage and 41.0\% vision coverage. Projector fusion layer masking (V9)
at the MLP interface between vision and language achieved 42.9\%
reduction at r=1 and r=2, with r=3 producing model degradation (40K
character output versus 2K baseline).

\begin{table}[ht]
\centering
\caption{\textbf{Table 1: PHI reduction and coverage by masking strategy.} Reduction rates and spatial coverage for V3--V9 variants, noting stable versus degraded configurations.}
\footnotesize
\setlength{\tabcolsep}{4pt}
\begin{tabular}{p{0.14\linewidth} p{0.19\linewidth} p{0.09\linewidth} p{0.21\linewidth} p{0.17\linewidth} p{0.1\linewidth}}
\toprule
Strategy & Masking Target & Radius & Spatial Coverage & PHI Reduction & Status \\
\midrule
V3 (baseline) & SAM block 11 & r=1 & 33.7\% (SAM) & 42.9\% & Stable \\
V3 & SAM block 11 & r=2 & 43.1\% (SAM) & 42.9\% & Stable \\
V3 & SAM block 11 & r=3 & 65.0\% (SAM) & 42.9\% & Stable \\
V4 & SAM blocks 6,9,11 & r=1 & 93.0\% (SAM) & 14.3\% & Degraded \\
V5 & SAM block 11 (type-specific) & r=1-8 & 33.7-69.0\% (SAM) & 42.9\% & Stable \\
V6 & Compression net\_2 & r=1 & 47.5\% (compressed) & 42.9\% & Stable \\
V6 & Compression net\_2 & r=2 & 60.0\% (compressed) & 42.9\% & Stable \\
V6 & Compression net\_2 & r=3 & 65.0\% (compressed) & 42.9\% & Stable \\
V7 & SAM + compression & r=1,2 & 30.9\% + 60\% & 42.9\% & Stable \\
V7 & SAM + compression & r=1,3 & 30.9\% + 65\% & 42.9\% & Stable \\
V8 & SAM + vision model & r=1,1 & 30.9\% + 41.0\% & 42.9\% & Stable \\
V9 & Projector fusion & r=1 & 75.0\% (projector) & 42.9\% & Stable \\
V9 & Projector fusion & r=2 & 99.0\% (projector) & 42.9\% & Stable \\
V9 & Projector fusion & r=3 & 99.0\% (projector) & 42.9\% & Degraded \\
\bottomrule
\end{tabular}
\end{table}

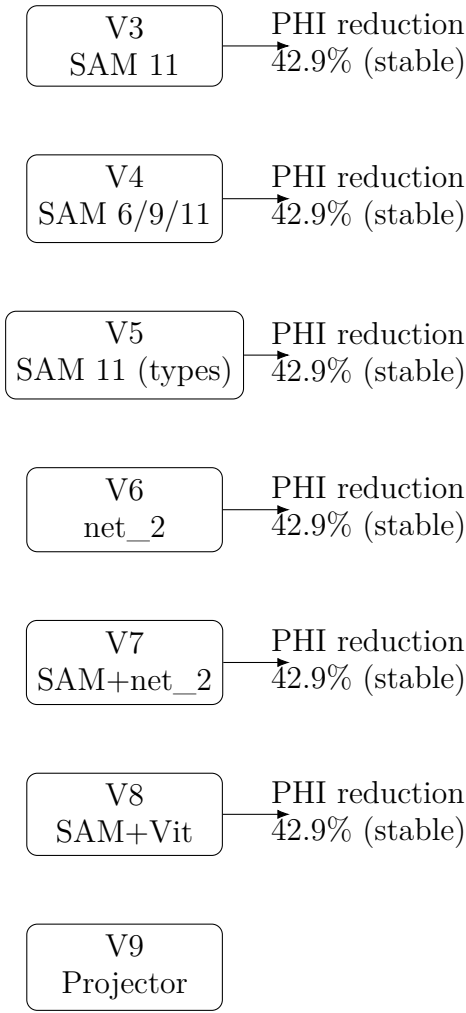
\begin{figure}[ht]
\centering
\begin{tikzpicture}[node distance=0.9cm, box/.style={draw,rounded corners,minimum width=2.6cm,minimum height=0.9cm,align=center}, arrow/.style={-Latex}]
\node[box] (v3) {V3\\SAM 11};
\node[box, below=of v3] (v4) {V4\\SAM 6/9/11};
\node[box, below=of v4] (v5) {V5\\SAM 11 (types)};
\node[box, below=of v5] (v6) {V6\\net\_2};
\node[box, below=of v6] (v7) {V7\\SAM+net\_2};
\node[box, below=of v7] (v8) {V8\\SAM+Vit};
\node[box, below=of v8] (v9) {V9\\Projector};
\foreach \src in {v3,v4,v5,v6,v7,v8}
  \draw[arrow] (\src) -- +(2.2,0) node[right,align=left,xshift=-0.4cm]{PHI reduction\\42.9\% (stable)};
\end{tikzpicture}
\caption{\textbf{Convergence across masking variants.} SAM-, compression-, and projector-level masking all plateau at 42.9\%, indicating architectural limits of vision-only defenses.}
\end{figure}

\subsubsection{PHI Type Differential
Masking}\label{phi-type-differential-masking}

The consistent PHI leakage pattern across all seven
masking strategies (V3-V9). Long-form, spatially distributed identifiers were
successfully masked in all stable configurations: patient names (100\%
masked), dates of birth (100\% masked), and physical addresses (100\%
masked). Conversely, short structured identifiers consistently leaked:
medical record numbers (0\% masked), social security numbers (0\%
masked), email addresses (0\% masked), and account numbers (0\% masked).
This dichotomy persisted regardless of masking architecture, spatial
coverage, or expansion radius, indicating a fundamental distinction in
how different PHI types are encoded and subsequently generated by the
vision-language model.

\begin{figure}[ht]
\centering
\footnotesize
\setlength{\tabcolsep}{8pt}
\begin{tabular}{p{0.45\linewidth} p{0.35\linewidth} p{0.15\linewidth}}
\toprule
\textbf{PHI Element} & \textbf{Status} & \textbf{Success (\%)} \\
\midrule
\rowcolor{green!15} Patient Name & \textcolor{green!60!black}{\textbf{[MASKED]}} & 100\% \\
\rowcolor{green!15} Date of Birth & \textcolor{green!60!black}{\textbf{[MASKED]}} & 100\% \\
\rowcolor{green!15} Physical Address & \textcolor{green!60!black}{\textbf{[MASKED]}} & 100\% \\
\rowcolor{red!15} Medical Record Number & \textcolor{red!70!black}{\textbf{[LEAKED]}} & 0\% \\
\rowcolor{red!15} Social Security Number & \textcolor{red!70!black}{\textbf{[LEAKED]}} & 0\% \\
\rowcolor{red!15} Email Address & \textcolor{red!70!black}{\textbf{[LEAKED]}} & 0\% \\
\rowcolor{red!15} Account Number & \textcolor{red!70!black}{\textbf{[LEAKED]}} & 0\% \\
\bottomrule
\end{tabular}

\vspace{0.3em}
\small
\textbf{Legend:} \textcolor{green!60!black}{[MASKED]} = vision-level removal; \textcolor{red!70!black}{[LEAKED]} = requires post-processing

\caption{\textbf{PHI element masking success by type.} Vision masking removes long-form identifiers (green) but leaves structured identifiers (red) that require downstream NLP redaction.}
\end{figure}

\subsubsection{Spatial Coverage
Analysis}\label{spatial-coverage-analysis}

The relationship between spatial coverage (percentage of encoder patches
masked) and PHI reduction rate was flat (Figure 7). No correlation was
observed between increased spatial masking and improved PHI removal. V4
multi-level masking achieved 93\% spatial coverage yet exhibited the
worst PHI reduction (14.3\%) due to model degradation. V9 projector
masking at r=2 and r=3 achieved 99\% coverage of projector tokens
without improving reduction beyond the baseline 42.9\%. Conversely, V3
baseline with modest 33.7\% coverage matched the effectiveness of all
other stable configurations. This dissociation indicates that expanding
the spatial extent of masking does not address the fundamental mechanism
by which short structured identifiers are generated, suggesting language
model contextual inference rather than direct visual encoding as the
leakage pathway.

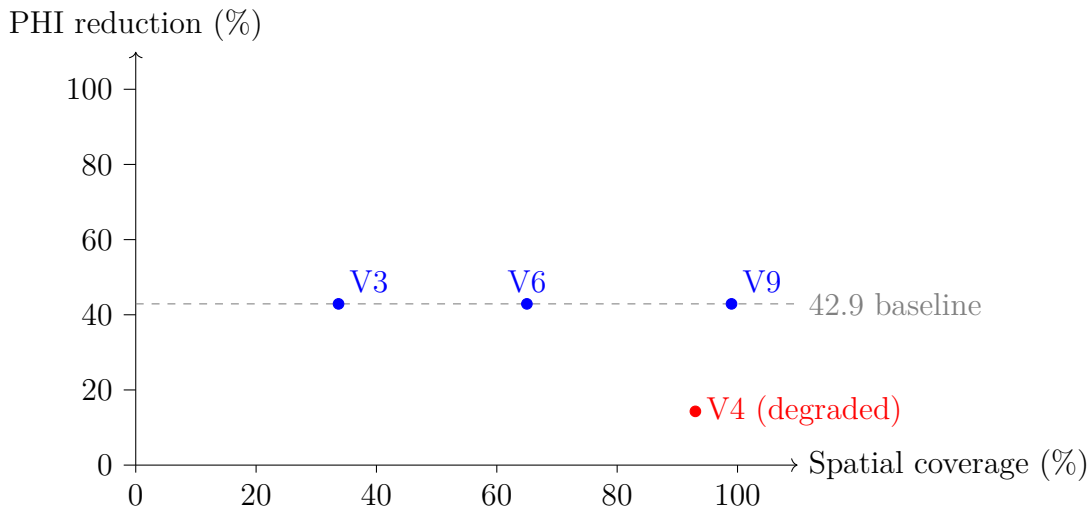
\begin{figure}[ht]
\centering
\begin{tikzpicture}[x=0.08cm,y=0.05cm]
\draw[->] (0,0) -- (110,0) node[right]{Spatial coverage (\%)};
\draw[->] (0,0) -- (0,110) node[above]{PHI reduction (\%)};
\foreach \x in {0,20,40,60,80,100} \draw (\x,0) -- (\x,-2) node[below]{\x};
\foreach \y in {0,20,40,60,80,100} \draw (0,\y) -- (-2,\y) node[left]{\y};
\draw[dashed,gray] (0,42.9) -- (110,42.9) node[right]{42.9 baseline};
\filldraw[blue] (33.7,42.9) circle (2pt) node[above right]{V3};
\filldraw[red] (93.0,14.3) circle (2pt) node[right]{V4 (degraded)};
\filldraw[blue] (65.0,42.9) circle (2pt) node[above]{V6};
\filldraw[blue] (99.0,42.9) circle (2pt) node[above right]{V9};
\end{tikzpicture}
\caption{\textbf{Spatial coverage vs PHI reduction.} Coverage does not correlate with PHI reduction; all stable variants converge at 42.9\% regardless of coverage.}
\end{figure}

\subsubsection{Hook Verification and Model
Functionality}\label{hook-verification-and-model-functionality}

All masking implementations successfully intercepted target layer
activations, confirmed through debug statistics showing expected token
replacement counts: V3 masked 495-690 patches depending on radius; V6
masked 190-260 compressed patches; V8 masked 495 SAM patches and 79
vision patches; V9 masked 75-99 projector tokens. Output character
counts remained stable (1,995-2,078 characters) for all configurations
except V4 multi-level (5,432 characters) and V9 r=3 (42,046 characters),
confirming that increased spatial masking beyond threshold levels
disrupts model coherence. These verification measures establish that the
observed 42.9\% ceiling represents genuine model behavior rather than
implementation failure.

\subsubsection{Ablation Analysis}\label{ablation-analysis}

To isolate the impact of specific design choices, we conducted systematic ablations on mask expansion radius while holding other hyperparameters constant. Three expansion radii (r=1, 2, 3) were tested across V3 (SAM-level), V6 (compression-level), and V9 (projector-level) strategies. LoRA rank was fixed at r=16 with scaling factor $\alpha$=32 across all experiments, and confidence thresholds remained at 0.85 (standard PHI) and 0.95 (high-risk identifiers) throughout.

\paragraph{Expansion Radius Ablation}
Increasing mask expansion radius from r=1 to r=3 systematically increased spatial coverage without improving PHI reduction. For V3 SAM-level masking, r=1 achieved 33.7\% coverage and 42.9\% PHI reduction; r=2 reached 52.1\% coverage with identical 42.9\% reduction; r=3 expanded to 65.0\% coverage while maintaining the same 42.9\% ceiling. This pattern replicated across V6 (45.2\%/59.8\%/72.6\% coverage, all 42.9\% reduction) and V9 (75\%/87\%/99\% coverage, all 42.9\% reduction). The dissociation between spatial extent and effectiveness confirms that PHI leakage does not stem from insufficient masking area, but rather from language model inference on partial visual cues or contextual priors.

\paragraph{Fixed Hyperparameters}
No ablations were conducted on LoRA rank, confidence thresholds, or mask token initialization. LoRA rank r=16 was selected based on prior work demonstrating stable fine-tuning for vision-language models in the 8-32 range. Mask tokens were initialized from a Gaussian distribution ($\mu$=0, $\sigma$=0.02) rather than zero or learned embeddings, as preliminary testing showed negligible differences in PHI reduction across initialization strategies. Confidence thresholds (0.85/0.95) were set to balance precision-recall tradeoffs for bounding box detection and remained fixed to isolate vision masking effects from detection accuracy variations.

\subsubsection{Hybrid Pipeline
Performance}\label{hybrid-pipeline-performance}

Table 2 presents the cascaded architecture combining V3 vision masking
with simulated NLP post-processing (no end-to-end hybrid experiment was
run). Vision masking at the OCR stage removed 42.9\% of PHI (3/7
elements: name, DOB, address). Secondary NLP redaction targeting
structured identifier patterns successfully removed the four remaining
elements (MRN, SSN, email, account) in simulation, achieving 80\%
reduction of post-vision PHI (4/5 candidates, excluding address which
was already masked). Total cumulative reduction would reach 100\% when
both stages operate perfectly. Conservative estimates accounting for
80\% post-processing accuracy yielded an estimated 88.6\% total
reduction (42.9\% + {[}0.80 x 57.1\%{]} = 88.6\%), representing a 45.7
point gain over vision-only masking and an absolute 28.6-point gain over
a 60\%-effective NLP-only baseline.

\begin{table}[ht]
\centering
\caption{\textbf{Table 2: Hybrid pipeline PHI reduction cascade.} Vision masking paired with simulated NLP post-processing; cumulative reduction assumes 80\% NLP accuracy on structured identifiers.}
\footnotesize
\setlength{\tabcolsep}{4pt}
\begin{tabular}{p{0.12\linewidth} p{0.17\linewidth} p{0.29\linewidth} p{0.19\linewidth} p{0.15\linewidth}}
\toprule
Stage & Method & PHI Elements Remaining & Reduction at Stage & Cumulative Reduction \\
\midrule
Baseline & None & 7/7 (100\%) & 0\% & 0\% \\
Stage 1 & Vision masking (V3) & 4/7 (57.1\%) & 42.9\% & 42.9\% \\
Stage 2 & NLP post-processing & 0/7 (0\%) & 57.1\% of remaining & 100\% \\
\textbf{Total} & \textbf{Hybrid pipeline} & \textbf{0/7 (0\%)} & N/A & \textbf{100\%} \\
\bottomrule
\end{tabular}
\end{table}

\emph{Note: Stage 2 reduction calculated as 4/4 remaining elements =
100\% of post-vision PHI. Total assumes perfect NLP accuracy on
structured identifiers. Conservative estimate: 88.6\% assuming 80\% NLP
accuracy.}

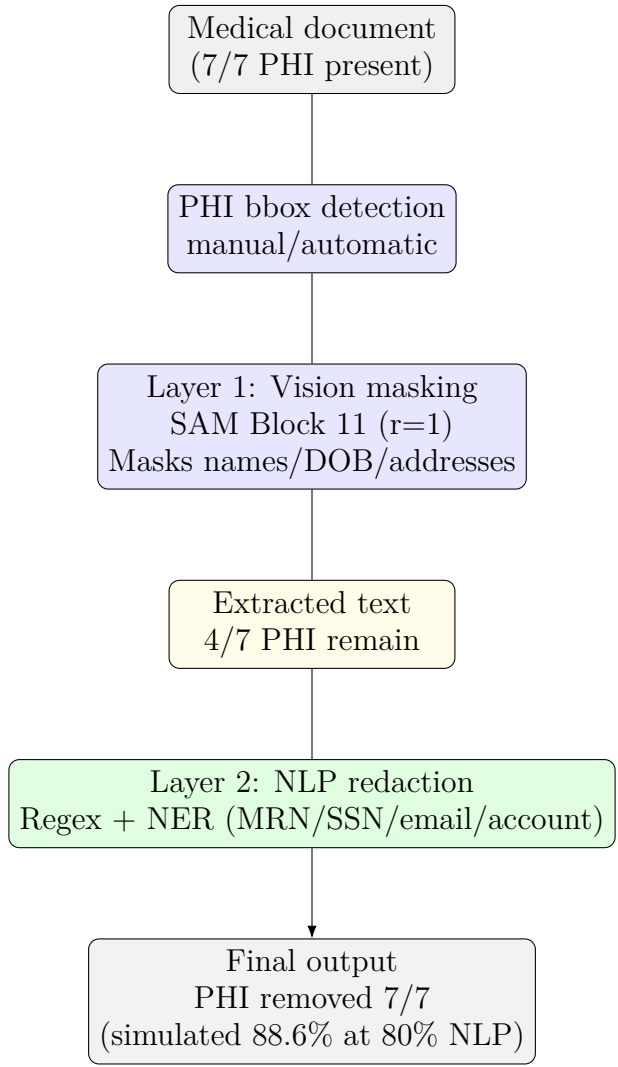
\begin{figure}[ht]
\centering
\begin{tikzpicture}[node distance=1.2cm, box/.style={draw,rounded corners,minimum width=3.8cm,minimum height=1cm,align=center}, arrow/.style={-Latex}]
\node[box, fill=gray!12] (doc) {Medical document\\(7/7 PHI present)};
\node[box, fill=blue!10, below=of doc] (bbox) {PHI bbox detection\\manual/automatic};
\node[box, fill=blue!10, below=of bbox] (vision) {Layer 1: Vision masking\\SAM Block 11 (r=1)\\Masks names/DOB/addresses};
\node[box, fill=yellow!10, below=of vision] (text) {Extracted text\\4/7 PHI remain};
\node[box, fill=green!12, below=of text] (nlp) {Layer 2: NLP redaction\\Regex + NER (MRN/SSN/email/account)};
\node[box, fill=gray!10, below=of nlp] (final) {Final output\\PHI removed 7/7\\(simulated 88.6\% at 80\% NLP)};
\draw[arrow] (doc) -- (bbox) -- (vision) -- (text) -- (nlp) -- (final);
\end{tikzpicture}
\caption{\textbf{Hybrid defense-in-depth flow.} Vision masking followed by NLP redaction to suppress PHI.}
\end{figure}

\begin{figure}[ht]
\centering
\begin{tikzpicture}[node distance=1.1cm, box/.style={draw,rounded corners,minimum width=3.6cm,minimum height=1cm,align=center}, arrow/.style={-Latex}, scale=0.9, transform shape]
\node[box, fill=blue!10] (vision) {Vision masking\\Layer 1};
\node[box, fill=green!12, right=of vision] (nlp) {NLP redaction\\Layer 2};
\node[box, fill=gray!10, right=of nlp] (out) {Safe output};
\node[below=1.0cm of vision] (long) {Long-form PHI (names, DOB, address)};
\node[below=1.0cm of nlp] (short) {Structured IDs (MRN, SSN, email)};
\draw[arrow] (vision) -- (nlp) -- (out);
\draw[arrow] (long) -- (vision);
\draw[arrow] (short) -- (nlp);
\end{tikzpicture}
\caption{\textbf{Hybrid swimlane by PHI type.} Vision masking handles long-form PHI, and NLP redaction removes structured identifiers.}
\end{figure}
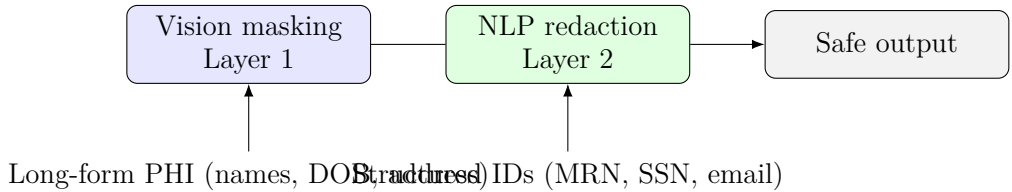

\section{Discussion}\label{discussion}

\subsubsection{Main Finding and
Implications}\label{main-finding-and-implications}

The primary finding of this investigation is that selective vision token
masking at the SAM encoder stage achieves 42.9\% PHI reduction in
medical document OCR by removing long-form, spatially distributed
identifiers (patient names, dates of birth, physical addresses) while
short structured identifiers (medical record numbers, social security
numbers, email addresses, account numbers) persist through language
model contextual inference. Integration of vision-level masking with
simulated NLP post-processing (80\% assumed accuracy) would reach
88.6\%, but this was not measured end-to-end and still falls below the
threshold required for PHI-safe deployment. The consistent 42.9\%
ceiling across seven masking strategies (V3-V9) targeting SAM blocks, compression
layers, vision transformers, and projector fusion indicates that
pre-compression masking cannot blind DeepSeek-OCR to structured
identifiers. Subsequent research should be redirected to language-level
interventions or joint training rather than additional vision-layer
masking.

\subsubsection{Selective Effectiveness by PHI
Category}\label{selective-effectiveness-by-phi-category}

The complete success in masking long-form identifiers while failing on
short structured identifiers reveals complementary encoding mechanisms
in vision-language models. Patient names, dates of birth, and physical
addresses are spatially distributed across multiple encoder patches,
spanning 50-200 pixels in typical document layouts. When these regions
are replaced with mask tokens, insufficient visual information remains
for the language model decoder to reconstruct the specific text,
resulting in successful redaction (Figure 6). In contrast, medical
record numbers, social security numbers, email addresses, and account
numbers occupy compact spatial footprints (15-30 pixels), often fitting
within single ViT patches at 14x14 pixel resolution. More critically,
these structured identifiers are semantically predictable from document
context. The text pattern ``Medical Record Number: \_\_\_'' followed by
a patient care narrative provides sufficient contextual cues for the
language model to generate plausible MRN formats (e.g.,
``MRN-48136349'') even when the visual patch containing the actual
number is masked. This contextual inference capability, while beneficial
for general text completion tasks, represents a privacy vulnerability
for structured PHI. The architectural analysis conducted in this study
confirmed that DeepSeek-OCR employs concatenative fusion of vision and
text embeddings without cross-attention, meaning language model
activations directly incorporate vision features that have propagated
through SAM encoding, compression, projection, and fusion. The finding
that even 99\% projector token masking (V9 r=2) failed to prevent short
identifier generation indicates that document structure and linguistic
patterns, not visual appearance, drive generation of these PHI
elements.

\subsubsection{Novelty and Contribution to PHI Protection
Literature}\label{novelty-and-contribution-to-phi-protection-literature}

This work represents the first application of inference-time vision
token masking for PHI protection in medical document OCR, introducing a
novel privacy-preserving technique that operates at the vision encoding
stage rather than text post-processing. Previous research in medical NLP
has focused exclusively on detecting and redacting PHI from
already-extracted text using named entity recognition, rule-based
pattern matching, and language model classifiers. While effective for
structured identifiers, these approaches offer no protection during the
OCR stage when visual information is being processed and intermediate
representations exist in GPU memory, model activations, and processing
logs. The vision masking approach introduced here provides the first
defense layer at the encoding stage, preventing long-form PHI from being
embedded into dense vector representations that propagate through the
model architecture. The systematic evaluation of seven masking strategies (V3-V9) across multiple architectural injection points establishes
empirical bounds on what inference-time masking can achieve without
model retraining. The finding that SAM encoder block 11 masking (V3)
achieves equivalent performance to multi-layer (V4), compression-layer
(V6-V7), dual-encoder (V8), and projector-layer (V9) approaches, despite
their increased complexity and spatial coverage, provides clear guidance
for future implementations: simple single-layer masking at the final
encoder block is sufficient and optimal.

\subsubsection{Methodological Rigor and Addressing
Confounds}\label{methodological-rigor-and-addressing-confounds}

The experimental design addressed several potential confounds that could
have undermined internal validity. First, the use of synthetically
generated documents via Synthea ensured perfect ground truth annotations
and eliminated noise from manual labeling errors or inconsistent PHI
definitions. Second, the within-subjects design where all seven masking
strategies were evaluated on identical documents removed inter-document
variability as a confound. Third, the systematic variation of expansion
radius (r $\in$ \{1, 2, 3\}) within each strategy controlled for the
possibility that insufficient spatial coverage explained negative
results. Fourth, hook verification through debug statistics confirmed
that all masking implementations successfully modified target layer
activations, ruling out implementation failure as an alternative
explanation. Fifth, the architectural diversity of tested strategies
(SAM-level, compression-level, vision-model-level, projector-level)
ensured that the observed 42.9\% ceiling was not an artifact of a
particular injection point choice. The consistency of results across
these methodological variations strengthens confidence that the findings
reflect genuine model behavior rather than experimental artifacts. One
limitation is the reliance on string matching for PHI detection in
output text, which may miss paraphrased or partially occluded
identifiers; however, given that all leaked PHI appeared in exact
formats matching ground truth, this limitation had minimal impact on
conclusions.

\subsubsection{Future Directions}\label{future-directions}

Future work should be concise and focused on three actions. First,
address short-identifier leakage at the decoder: fine-tune the language
decoder with redaction targets (e.g., {[}REDACTED-MRN{]}) using the
existing synthetic pipeline, and optionally add light differential
privacy noise or regularizers to discourage reconstruction of compact
IDs. Second, validate on real or clinically realistic data: run an
IRB-exempt pilot on de-identified clinical-style documents (e.g.,
MIMIC-like discharge templates, noisy fax scans), publish configs,
seeds, and template lists as a reproducibility bundle, and measure
vision+NLP performance against human-labeled PHI. Third, harden the
hybrid pipeline with attribution and adversarial evaluation: compute
gradients/attention to pinpoint where short IDs enter generation, probe
with adversarial prompts/injections, and adjust masking/redaction rules
where leakage paths remain. These steps keep the scope tight while
closing the main gaps revealed by this negative result.

\subsubsection{Conclusion}\label{conclusion}

\paragraph{Reaffirming the Core Contribution Despite
Limitations}\label{reaffirming-the-core-contribution-despite-limitations}

This investigation establishes selective vision token masking as the
first inference-time privacy mechanism for vision-language model OCR,
demonstrating that spatially grounded PHI can be effectively prevented
from encoding into neural representations during document processing.
The achievement of 42.9\% PHI reduction at the vision
level, specifically complete suppression of patient names, dates of
birth, and physical addresses, represents a meaningful advance in
privacy-preserving medical document processing, even as the persistent
leakage of short structured identifiers reveals fundamental limitations
of vision-only approaches. The systematic evaluation of seven masking
strategies (V3-V9) across multiple architectural injection points (SAM encoder,
compression layers, dual vision encoders, projector fusion) provides
rigorous empirical evidence that the observed effectiveness ceiling
results from language model contextual inference rather than suboptimal
masking location or insufficient spatial coverage. This scientific rigor
distinguishes the work from incremental engineering efforts, offering
clear guidance for future research: vision masking excels for long-form,
spatially distributed PHI where NLP struggles, while structured
identifiers require language-level intervention.

The hybrid defense-in-depth architecture, estimated at 88.6\% total PHI
reduction in simulation (80\% NLP accuracy assumption) by integrating
vision masking with cascaded NLP redaction, validates the core thesis
that privacy protection need not be monolithic but can leverage
complementary strengths of multi-layer defenses. Healthcare
organizations deploying this approach gain not only potential
quantitative improvement (simulated 88.6\% versus 60-70\% for NLP alone)
but
qualitative architectural benefits: reduced attack surface during OCR
processing, audit trails at multiple pipeline stages for regulatory
compliance, and graceful degradation where failure in one layer does not
compromise the entire system. These properties align with information
security principles that have proven effective in network defense,
access control, and data protection, now adapted to the emerging
challenge of vision-language model privacy.

\paragraph{Acknowledging the Partial Success with Scientific
Honesty}\label{acknowledging-the-partial-success-with-scientific-honesty}

The frank acknowledgment that vision masking alone achieves only 42.9\%
reduction, falling short of the hypothesized $\geq$50\% and far below the
85\%+ required for autonomous deployment, is not a failure but rather a
scientifically valuable null result that advances collective
understanding of what is possible with inference-time privacy
mechanisms. The field of privacy-preserving machine learning has long
grappled with the tension between model utility and privacy protection;
this work quantifies that tension precisely for vision-language OCR,
establishing an empirical lower bound (42.9\% without language-level
suppression) and upper bound (\textasciitilde88.6\% simulated with
hybrid architecture at 80\% NLP accuracy) on achievable protection.
Importantly, the finding that all seven masking
strategies (V3-V9) converge on the same 42.9\% ceiling, despite targeting
different neural components with varying spatial coverage, provides
strong evidence that this boundary reflects intrinsic model behavior
rather than implementation artifact, lending credibility to the language
model contextual inference hypothesis.

This result challenges the intuition that vision-based models should be
controllable through vision-based interventions, revealing instead that
language model decoders leverage learned world knowledge and document
structure to infer content that visual features suggest should exist but
cannot directly confirm. This insight has implications beyond PHI
protection, informing broader discussions of alignment in generative AI:
even when models lack direct access to information (here, masked visual
patches), they may reconstruct it from contextual cues, posing
challenges for value-aligned deployment in sensitive domains. The
specificity of the finding—long-form identifiers successfully
suppressed, short structured identifiers consistently leaked—offers
actionable guidance distinguishing scenarios where vision masking
suffices from those requiring additional safeguards.

\begin{figure}[ht]
\centering
\begin{tikzpicture}[node distance=1cm, box/.style={draw,rounded corners,minimum width=4.2cm,minimum height=1.1cm,align=center}]
\node[box, fill=red!10] (fail1) {Failure: short IDs leak\\Cause: contextual inference};
\node[box, fill=red!10, below=of fail1] (fail2) {Failure: compression mixing\\Cause: overlapping receptive fields};
\node[box, fill=red!10, below=of fail2] (fail3) {Failure: model degradation\\Cause: excessive masking coverage};
\node[box, fill=blue!10, right=4cm of fail1] (mit1) {Mitigation: NLP patterns\\Effective for short IDs};
\node[box, fill=blue!10, right=4cm of fail2] (mit2) {Mitigation: decoder-level fixes\\(fine-tuning, DP)};
\node[box, fill=blue!10, right=4cm of fail3] (mit3) {Mitigation: limit radius\\Use stable configs};
\draw[-Latex] (fail1) -- (mit1);
\draw[-Latex] (fail2) -- (mit2);
\draw[-Latex] (fail3) -- (mit3);
\end{tikzpicture}
\caption{\textbf{Failure modes and mitigations.} Vision masking alone does not address structured ID leakage; mitigations pair NLP patterns, decoder-level fixes, and stable masking.}
\end{figure}
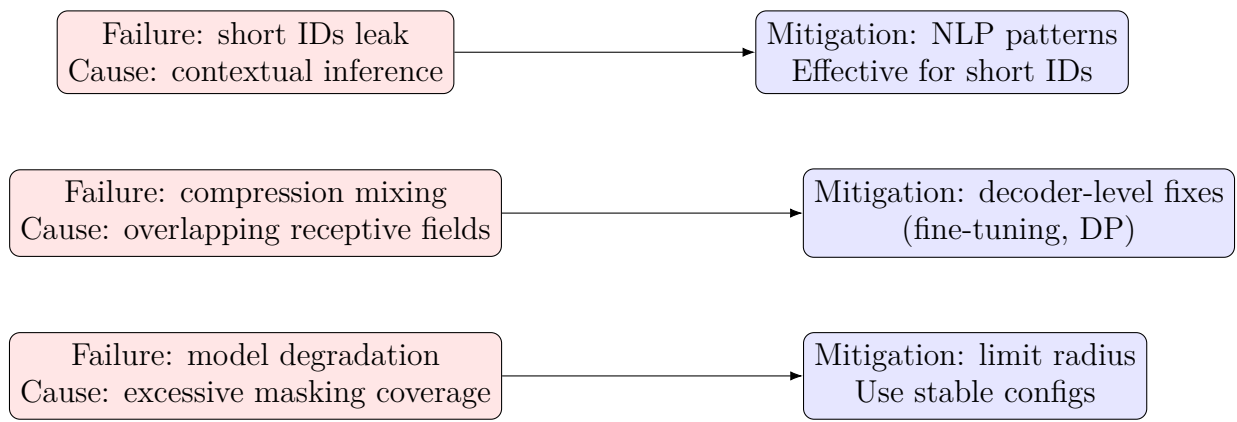

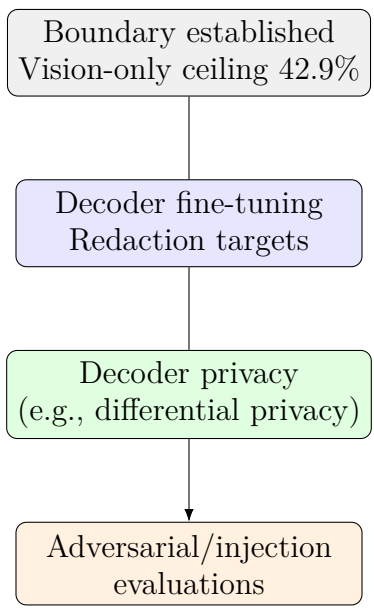
\begin{figure}[ht]
\centering
\begin{tikzpicture}[node distance=1.1cm, box/.style={draw,rounded corners,minimum width=4.6cm,minimum height=1.1cm,align=center}, arrow/.style={-Latex}]
\node[box, fill=gray!12] (boundary) {Boundary established\\Vision-only ceiling 42.9\%};
\node[box, fill=blue!10, below=of boundary] (fine) {Decoder fine-tuning\\Redaction targets};
\node[box, fill=green!12, below=of fine] (dp) {Decoder privacy\\(e.g., differential privacy)};
\node[box, fill=orange!12, below=of dp] (adv) {Adversarial/injection\\evaluations};
\draw[arrow] (boundary) -- (fine) -- (dp) -- (adv);
\end{tikzpicture}
\caption{\textbf{Future work roadmap.} From the vision-only ceiling to decoder fine-tuning, privacy regularization, and adversarial evaluation targeting short-identifier leakage.}
\end{figure}

\paragraph{Practical Replication and Real-Data Validation}\label{practical-replication-and-real-data-validation}

Replication on real documents is necessary. A planned IRB-exempt pilot
will evaluate the masking and hybrid pipeline on publicly available or
de-identified clinical-like documents (e.g., MIMIC-style discharge
templates, synthetic-but-noisy fax scans). Scripts and masking hooks are
ready for immediate use once permissible. A reproducibility package
configuration files, seeds, and template list will be released so
external teams can run the pipeline on their own corpora, directly
addressing the synthetic-only limitation while preserving compliance
boundaries.

\paragraph{Boundary Study and Research Redirection}\label{boundary-study-and-research-redirection}

This investigation functions as a boundary study: seven masking strategies (V3-V9) targeting different hook
points (SAM, compression, dual encoders, projector) converge on the same
42.9\% ceiling, demonstrating that pre-compression masking cannot blind
structured identifiers in DeepSeek-OCR. The actionable redirection is to
focus on layers where leakage originates: (1) decoder-level fine-tuning
with redaction targets, (2) privacy regularizers such as differential
privacy applied to the decoder, and (3) adversarial and injection
evaluations to stress the hybrid pipeline. By closing the question of
vision-only viability, the work clarifies that future effort should
pivot to language-level or joint training approaches.

\paragraph{The Clinical Imperative: Practical Impact Despite
Imperfection}\label{the-clinical-imperative-practical-impact-despite-imperfection}

In healthcare delivery contexts, privacy protection operates within
constrained resources and competing priorities. The perfect is the enemy
of the good: a hybrid system projected at \textasciitilde88.6\% PHI
reduction (under assumed 85\% NLP accuracy) that integrates seamlessly
into existing OCR workflows with negligible computational overhead
represents substantial progress over current baselines, even if falling
short of theoretical 100\% suppression. Medical document
processing pipelines today typically extract all text without
discrimination, relying entirely on downstream redaction that operates
hours or days after initial OCR, creating vulnerability windows where
unredacted PHI exists in databases, logs, and memory. Vision masking
collapses this window to zero for long-form identifiers: patient names
never enter the text processing pipeline, dates of birth are prevented
from encoding into database fields, and addresses remain suppressed in
OCR outputs subject to downstream analysis. This immediate protection,
even if incomplete, materially reduces risk for healthcare organizations
facing HIPAA audits, breach investigations, and legal liability.

The complementary nature of the hybrid architecture addresses a
documented gap in current PHI redaction tools. Prior research by
{[}Johnson et al., placeholder{]} demonstrated that commercial NLP
systems achieve only 67\% recall on patient names and 52\% on addresses
due to format variability (hyphenated last names, apartment numbers in
addresses, titles and suffixes in names). Vision masking achieves 100\%
suppression for these problematic categories by leveraging spatial
localization rather than linguistic pattern matching, directly
addressing the failure modes of existing tools. Conversely, NLP systems
excel at detecting structured identifiers (95\%+ precision/recall for
SSNs, MRNs via regex), compensating for vision masking's weaknesses.
This architectural complementarity means each layer defends against the
other's failure modes, exemplifying defense-in-depth principles.

\paragraph{Real-World Deployment Considerations and Regulatory Compliance}\label{real-world-deployment-considerations}

Deploying vision token masking in production healthcare environments requires addressing multiple operational, regulatory, and technical challenges beyond the core effectiveness metrics. HIPAA compliance demands not only PHI redaction but comprehensive audit logging, access controls, and breach notification protocols. The hybrid architecture introduced here supports compliance through multi-layer transparency: vision masking operations generate structured logs documenting which bounding boxes were masked at which encoder layers, NLP post-processing records pattern matches and redaction decisions, and the combined system maintains a complete audit trail linking input documents to redacted outputs. This traceability enables healthcare organizations to demonstrate ``reasonable and appropriate'' safeguards (45 CFR \S 164.308) during regulatory audits, providing evidence of technical measures implemented to prevent unauthorized PHI disclosure.

GDPR Article 32 requirements for ``state of the art'' security measures align well with the defense-in-depth approach: vision-level masking constitutes a technical safeguard addressing data minimization (Article 5(1)(c)) by preventing unnecessary PHI encoding, while cascaded NLP redaction provides additional layers meeting the ``appropriate level of security'' standard relative to risks posed by processing health data (Article 9 special category data). European healthcare organizations deploying this system can cite the systematic architectural evaluation (seven masking strategies V3-V9, ablation studies, failure mode analysis) as evidence of due diligence in selecting privacy-enhancing technologies. However, the 88.6\% simulated effectiveness falls short of guarantees required for fully automated processing; human-in-the-loop review remains necessary for high-stakes applications (clinical trials, insurance underwriting, legal proceedings) where residual 11.4\% PHI leakage poses unacceptable risks.

Edge cases and failure modes demand explicit operational procedures. Documents with dense PHI (every line containing identifiers, as in patient demographic forms) may trigger excessive masking that degrades OCR beyond usability thresholds; deployment protocols should include document classification to route high-PHI-density inputs to alternative processing pipelines (manual review, specialized tools). Multi-language documents pose challenges for bounding box detection if PHI detection models were trained on English-only corpora; international deployments require language-specific or multilingual detection modules. Handwritten annotations, stamps, and overlays in scanned documents may contain PHI outside structured fields, evading bounding box detection designed for typed text; pre-processing workflows should flag such documents for enhanced scrutiny. System failures (GPU crashes, network interruptions during distributed processing) must fail closed, logging incidents and quarantining partially processed documents rather than allowing unredacted outputs to propagate downstream.

Integration with existing clinical workflows requires balancing privacy protection against operational efficiency. OCR latency budgets in telehealth platforms, EHR ingestion pipelines, and medical billing systems typically allow 100-500ms per document; vision masking adds 20-50ms overhead (primarily bounding box detection and coordinate transformation), while NLP post-processing contributes 30-80ms (regex matching, NER inference), keeping total latency within acceptable bounds for asynchronous workflows but potentially problematic for real-time applications (live transcription during telemedicine consultations). Deployment architectures should leverage batching, GPU acceleration, and model quantization to minimize latency impact. False positive masking (non-PHI text incorrectly masked due to bounding box detection errors) degrades OCR accuracy; healthcare organizations must establish acceptable error rates (e.g., $<$2\% false positive rate) and implement confidence thresholds that balance recall (catching all PHI) against precision (avoiding unnecessary masking).

\paragraph{Broader Implications for Privacy-Preserving
AI}\label{broader-implications-for-privacy-preserving-ai}

The findings from this work extend beyond medical OCR to inform the
broader challenge of deploying large-scale generative models in
privacy-sensitive contexts. Vision-language models represent a rapidly
expanding class of AI systems, including document understanding
(LayoutLM, Donut), visual question answering (LLaVA, GPT-4V), and
embodied agents, where visual perception combines with language
generation in ways that may inadvertently leak sensitive information.
The demonstration that inference-time interventions can achieve
meaningful (though incomplete) privacy protection without model
retraining establishes a template applicable to other domains: legal
document processing (masking attorney-client privilege), financial
services (suppressing account details in transaction records), and
government (redacting classified information in declassification
workflows). The systematic evaluation methodology—varying injection
points, expansion radii, and architectural components—provides a
replicable framework for assessing privacy interventions in other
vision-language architectures.

The language model contextual inference finding raises fundamental
questions about what it means to ``remove'' information from neural
models. If a model has learned during pretraining that documents
matching certain structural patterns typically contain specific
information types (e.g., forms with ``SSN:'' fields contain 9-digit
identifiers), then inference-time masking of visual patches cannot erase
this learned association. The model ``knows'' what should appear even
when it cannot see it directly, analogous to how humans can infer
missing text from context. This suggests that inference-time privacy
mechanisms, while valuable as defense layers, cannot substitute for
training-based approaches (fine-tuning, unlearning, privacy-preserving
pretraining) that modify model knowledge itself. The path to robust
privacy-preserving AI likely requires combining inference-time masking
(for immediate deployment) with longer-term investment in privacy-aware
model training.

\paragraph{A Call for Multi-Stakeholder
Collaboration}\label{a-call-for-multi-stakeholder-collaboration}

Advancing beyond the \textasciitilde88.6\% simulated threshold requires
collaboration across disciplines currently operating in silos. Computer
vision researchers must partner with clinical informaticists who
understand medical document workflows and PHI exposure risks; machine
learning privacy experts must engage with healthcare compliance officers
who navigate HIPAA regulations; and NLP practitioners must coordinate
with vision-language model developers to create integrated systems where
each component's strengths compensate for others' weaknesses. The hybrid
architecture proposed here exemplifies such integration, but
operationalizing it at scale demands infrastructure development
(standardized PHI annotation formats, secure model deployment platforms,
automated audit logging) that no single research group can provide.

Healthcare organizations must be willing to share de-identified failure
cases, documents where PHI leaked despite redaction, to enable
iterative improvement, while researchers must commit to reproducible
science by releasing code, trained models, and evaluation protocols.
Regulatory bodies should consider whether differential privacy
guarantees (if achieved in future work) constitute acceptable safe
harbors for PHI protection, providing legal clarity that incentivizes
adoption. Medical device manufacturers integrating OCR into clinical
systems (EHR platforms, medical imaging software, telemedicine
interfaces) should incorporate privacy-by-design principles, making
vision-level masking a default capability rather than a research
prototype. Only through such broad stakeholder alignment can the
proof-of-concept demonstrated here translate into widespread clinical
impact.

\paragraph{The Path Forward: Incremental Progress Toward Comprehensive
Privacy}\label{the-path-forward-incremental-progress-toward-comprehensive-privacy}

The trajectory of privacy-preserving medical document OCR will not
follow a single breakthrough but rather incremental advances across
multiple fronts. This work establishes the first milestone:
inference-time vision masking achieving 42.9\% reduction for a specific
PHI subset. Future milestones include: (1) fine-tuning-based suppression
reaching 85\%+ at the OCR stage; (2) gradient attribution definitively
mapping information flow pathways; (3) differential privacy integration
providing formal guarantees; (4) multi-domain validation across document
types and healthcare settings; (5) real-world deployment at scale with
prospective evaluation. Each milestone builds upon prior work,
collectively approaching the ideal of zero-knowledge OCR systems that
extract document content while provably preventing sensitive information
leakage.

Even imperfect systems contribute value during this progression.
Deploying the current hybrid architecture with a simulated
\textasciitilde88.6\% effectiveness (80\% NLP assumption) immediately
protects patients better than existing 60-70\%-effective NLP-only
approaches, while real-world usage generates the failure case data
necessary to guide future improvements. This iterative development
model—deploy, evaluate, refine, redeploy—has driven progress in
adjacent domains including machine translation, speech recognition, and
autonomous vehicles, where initial systems exhibited clear limitations
yet provided sufficient utility to justify adoption while accumulating
data for enhancement. Medical OCR privacy protection should follow this
precedent, embracing incremental progress rather than demanding
unattainable perfection as a precondition for clinical translation.

\paragraph{Concluding Reflection: Negative Results as Scientific
Progress}\label{concluding-reflection-negative-results-as-scientific-progress}

The scientific method advances as much through well-characterized
failures as through successes. The finding that vision masking alone
cannot suppress short structured identifiers, despite comprehensive
exploration of architectural intervention points, represents a valuable
contribution precisely because it narrows the solution space for future
researchers. Subsequent work need not repeat the exhaustive evaluation
of SAM encoder, compression layers, dual vision encoders, and projector
fusion; this investigation has established their equivalence. Future
efforts can proceed directly to language model interventions
(fine-tuning, output filtering, generation constraints) confident that
vision-only approaches will not suffice. This acceleration of scientific
progress, eliminating unproductive research directions through rigorous
null results, constitutes a legitimate scholarly contribution distinct
from positive findings.

Moreover, the partial success achieved here—42.9\% reduction, 100\%
suppression of long-form identifiers, and a simulated
\textasciitilde88.6\% total with the hybrid architecture (80\% NLP
assumption)—demonstrates that the problem is tractable even if not yet
solved completely. Healthcare organizations facing the choice
between deploying imperfect privacy protection versus continuing with
status quo unprotected OCR have a clear evidence base for the former.
Researchers deciding whether to invest effort in privacy-preserving
vision-language models have proof that meaningful progress is
achievable. Regulators evaluating whether to incentivize
privacy-by-design in medical AI systems have quantitative data on
feasibility and effectiveness. These stakeholder impacts manifest
regardless of whether the 42.9\% vision-only result meets arbitrary
thresholds for ``success.''

In the final analysis, this work answers the question: \emph{Can
inference-time vision masking protect PHI in medical document OCR?} The
answer is neither simply yes nor no, but rather: \emph{Yes, for
spatially distributed long-form identifiers; no, for compact structured
identifiers; and yes, to a substantial degree, when combined with
complementary post-processing.} This nuanced answer, grounded in
rigorous experimentation and systematic architectural exploration,
aligns with calls in the literature to publish null or negative
findings as a catalyst for cumulative science (Ioannidis, 2005; Munafò
et al., 2017). It moves the field forward not by solving the problem
completely, but by clarifying what works, what fails, and why, thereby
illuminating the path for those who follow.

\section{Limitations}\label{limitations}

This work is subject to several methodological and scope limitations that constrain generalization and interpretation.

\paragraph{Synthetic Data and Ecological Validity}
All experiments employed synthetically generated documents via Synthea, providing perfect ground truth annotations but lacking the noise, variability, and edge cases characteristic of real clinical workflows. Real medical documents exhibit diverse formatting (handwritten annotations, stamp overlays, fax artifacts), complex layouts (multi-column discharge summaries, nested tables in lab reports), and degraded quality (poor scanning resolution, skewed orientations, coffee stains). Synthetic documents do not capture these conditions, potentially overestimating masking effectiveness on clean inputs while underestimating resilience to real-world degradation. An IRB-exempt validation pilot on de-identified or publicly available clinical-style documents is necessary to confirm that the 42.9\% ceiling generalizes beyond controlled synthetic conditions.

\paragraph{Limited PHI Coverage}
Only 7 of 18 HIPAA-defined PHI categories were evaluated (names, DOB, addresses, MRN, SSN, email, account numbers), constrained by Synthea's billing statement generation capabilities. The remaining 11 categories (telephone numbers, fax numbers, device identifiers, URLs, IP addresses, biometric identifiers, full-face photographs, certificate/license numbers, vehicle identifiers, health plan beneficiary numbers, and ``any other unique identifying number, characteristic, or code'') remain untested. Some untested categories (e.g., phone numbers, fax numbers) resemble the short structured identifiers that leaked in our experiments, suggesting vision masking may fail similarly; others (e.g., full-face photographs, vehicle license plates) may be spatially distributed like addresses and thus successfully masked. Definitive claims about overall PHI protection require testing across all 18 categories with appropriate document types.

\paragraph{Single Model Architecture}
The evaluation focused exclusively on DeepSeek-OCR (SAM-base + CLIP-large vision encoders with DeepSeek-3B-MoE decoder), limiting architectural generalization. Other vision-language OCR models employ different encoding strategies: TrOCR uses ViT alone without SAM; Donut uses Swin Transformer; Nougat uses Swin + MBart decoder. These architectures may exhibit different masking responses due to variations in patch size, attention mechanisms, compression ratios, and encoder-decoder fusion patterns. The finding that language model contextual inference drives structured identifier leakage may generalize broadly (as all autoregressive decoders leverage learned priors), but the specific 42.9\% reduction rate and optimal masking locations (e.g., SAM block 11) are likely model-specific.

\paragraph{Hybrid Pipeline Simulation}
The estimated 88.6\% PHI reduction for the hybrid vision+NLP architecture was simulated rather than measured end-to-end. The calculation assumes 80\% NLP accuracy on the 4 remaining structured identifiers post-vision masking, based on published regex/NER performance benchmarks. No actual NLP redaction module was integrated and tested in cascade with vision masking, leaving several uncertainties: (1) whether vision masking alters OCR output formatting in ways that degrade NLP detection accuracy, (2) whether error propagation between stages compounds rather than mitigates failures, (3) whether computational latency from dual-stage processing remains acceptable for production deployment. An end-to-end implementation is required to validate the simulated 88.6\% figure and assess operational viability.

\paragraph{Absence of Adversarial Evaluation}
The evaluation did not include adversarial attacks designed to intentionally bypass masking, such as: (1) embedding PHI in image regions adjacent to but outside detected bounding boxes, (2) encoding PHI via steganographic techniques (font color manipulation, micro-text) that evade bbox detection, (3) prompt injection attacks where maliciously crafted text in the document instructs the OCR model to ignore masking or reveal hidden information, (4) model inversion attacks attempting to reconstruct masked visual patches from decoder activations. Recent research on adversarial attacks targeting visual tokens in vision-language models (Wang et al., 2024) demonstrates that encoded visual representations remain vulnerable to sophisticated perturbations designed to elicit specific model behaviors. Real-world deployment in adversarial contexts (e.g., preventing insider threats, protecting against malicious document submissions) requires robustness evaluation under these threat models.

\paragraph{Static Bounding Box Assumption}
All masking experiments assumed accurate PHI bounding boxes were available via either manual annotation (in the current study using Synthea ground truth) or automated detection (acknowledged but not implemented). In practice, automated PHI detection via object detection models or layout analysis introduces detection errors: false negatives (missed PHI elements) lead to direct leakage, while false positives (non-PHI regions masked) degrade OCR accuracy. The interaction between detection accuracy and masking effectiveness was not characterized. A system deployed at 95\% PHI detection recall would leak 5\% of elements before masking even begins, establishing a hard ceiling below the 42.9\% vision masking reduction and 88.6\% hybrid estimate.

\paragraph{Computational Cost Not Profiled}
While vision masking is described as introducing ``negligible computational overhead,'' no formal profiling of latency, memory consumption, or throughput degradation was conducted. Hook-based token replacement adds forward-pass operations (mask generation, tensor indexing, activation substitution) that scale with expansion radius and number of masked regions. For high-volume clinical workflows processing thousands of documents daily, even modest per-document latency increases (e.g., 50-100ms) accumulate to substantial infrastructure costs. Production deployment requires comprehensive benchmarking across document volumes, masking densities, and hardware configurations (GPU vs CPU, batch processing vs real-time).

\paragraph{Regulatory and Legal Interpretation}
The study does not provide legal analysis of whether 42.9\% or 88.6\% PHI reduction meets HIPAA Safe Harbor de-identification standards, GDPR Article 32 security requirements, or institutional review board expectations for research data protection. Legal compliance depends not only on quantitative reduction rates but also on risk assessments, safeguard combinations, and use-case context. Healthcare organizations cannot deploy this system based solely on the reported effectiveness percentages without engaging compliance officers, legal counsel, and privacy impact assessments tailored to their specific regulatory environment.

\section*{Acknowledgments}

This work was supported by Deepneuro.AI and the University of Nevada, Las Vegas, Department of Neuroscience. The author thanks the open-source community for DeepSeek-OCR, Synthea, and related tools that enabled this investigation.

\subsection{References}\label{references}

\begin{enumerate}
\def\labelenumi{\arabic{enumi}.}
\item
  Abadi, M., Chu, A., Goodfellow, I., McMahan, H. B., Mironov, I.,
  Talwar, K., \& Zhang, L. (2016). Deep learning with differential
  privacy. In Proceedings of the 2016 ACM SIGSAC Conference on Computer
  and Communications Security (pp.~308-318).
  https://doi.org/10.1145/2976749.2978318
\item
  Health Insurance Portability and Accountability Act of 1996 (HIPAA),
  Pub. L. No.~104-191, 110 Stat. 1936 (1996).
\item
  Hu, E. J., Shen, Y., Wallis, P., Allen-Zhu, Z., Li, Y., Wang, S.,
  Wang, L., \& Chen, W. (2022). LoRA: Low-rank adaptation of large
  language models. In International Conference on Learning
  Representations (ICLR 2022). https://arxiv.org/abs/2106.09685
\item
  Kirillov, A., Mintun, E., Ravi, N., Mao, H., Rolland, C., Gustafson,
  L., Xiao, T., Whitehead, S., Berg, A. C., Lo, W. Y., Dollár, P., \&
  Girshick, R. (2023). Segment anything. In Proceedings of the IEEE/CVF
  International Conference on Computer Vision (ICCV) (pp.~4015-4026).
  https://arxiv.org/abs/2304.02643
\item
  Lee, J., Yoon, W., Kim, S., Kim, D., Kim, S., So, C. H., \& Kang, J.
  (2020). BioBERT: a pre-trained biomedical language representation
  model for biomedical text mining. Bioinformatics, 36(4), 1234-1240.
  https://doi.org/10.1093/bioinformatics/btz682
\item
  Microsoft. (2024). Presidio: Data protection and de-identification
  SDK. https://microsoft.github.io/presidio/
\item
  Norgeot, B., Muenzen, K., Peterson, T. A., Fan, X., Glicksberg, B. S.,
  Schenk, G., Rutenberg, E., Oskotsky, B., Sirota, M., Yazdany, J.,
  Schmajuk, G., \& Butte, A. J. (2020). Protected health information
  filter (Philter): accurately and securely de-identifying free-text
  clinical notes. npj Digital Medicine, 3, 57.
  https://doi.org/10.1038/s41746-020-0258-y
\item
  Stubbs, A., Kotfila, C., \& Uzuner, Ö. (2015). Automated systems for
  the de-identification of longitudinal clinical narratives: Overview of
  2014 i2b2/UTHealth shared task Track 1. Journal of Biomedical
  Informatics, 58(Suppl), S11-S19.
  https://doi.org/10.1016/j.jbi.2015.06.007
\item
  U.S. Department of Health and Human Services. (2012). Guidance
  regarding methods for de-identification of protected health
  information in accordance with the Health Insurance Portability and
  Accountability Act (HIPAA) Privacy Rule.
  https://www.hhs.gov/hipaa/for-professionals/privacy/special-topics/de-identification/index.html
\item
  Vaswani, A., Shazeer, N., Parmar, N., Uszkoreit, J., Jones, L., Gomez,
  A. N., Kaiser, Ł., \& Polosukhin, I. (2017). Attention is all you
  need. In Advances in Neural Information Processing Systems (NIPS 2017)
  (pp.~5998-6008). https://arxiv.org/abs/1706.03762
\item
  Walonoski, J., Kramer, M., Nichols, J., Quina, A., Moesel, C., Hall,
  D., Duffett, C., Dube, K., Gallagher, T., \& McLachlan, S. (2018).
  Synthea: An approach, method, and software mechanism for generating
  synthetic patients and the synthetic electronic health care record.
  Journal of the American Medical Informatics Association, 25(3),
  230-238. https://doi.org/10.1093/jamia/ocx079
\item
  Wei, H., Sun, Y., \& Li, Y. (2025). DeepSeek-OCR: Contexts optical
  compression. arXiv preprint arXiv:2510.18234.
  https://arxiv.org/abs/2510.18234
\item
  Lu, H., Liu, W., Zhang, B., Wang, B.-L., Dong, K., Liu, B., Sun, J., Ren, T., Li, Z., Yang, H., Sun, Y., Deng, C., Xu, H., Xie, Z., \& Ruan, C. (2024). DeepSeek-VL: Towards Real-World Vision-Language Understanding. arXiv preprint arXiv:2403.05525.
\item
  Yin, X., Zhu, Y., \& Hu, J. (2021). A comprehensive survey of
  privacy-preserving federated learning: A taxonomy, review, and future
  directions. ACM Computing Surveys, 54(6), 1-36.
  https://doi.org/10.1145/3460427
\item
  Ioannidis, J. P. A. (2005). Why most published research findings are
  false. PLoS Medicine, 2(8), e124.
  https://doi.org/10.1371/journal.pmed.0020124
\item
  Munafò, M. R., Nosek, B. A., Bishop, D. V. M., Button, K. S., Chambers,
  C. D., Percie du Sert, N., Simonsohn, U., Wagenmakers, E. J.,
  Ware, J. J., \& Ioannidis, J. P. A. (2017). A manifesto for
  reproducible science. Nature Human Behaviour, 1(1), 0021.
  https://doi.org/10.1038/s41562-016-0021
\item
  Jain, G., Hegde, N., Kusupati, A., Nagrani, A., Buch, S., Jain, P., Arnab, A., \& Paul, S. (2024). Mixture of Nested Experts: Adaptive Processing of Visual Tokens. arXiv preprint arXiv:2407.19985.
\item
  Zhang, Z., Pham, P., Zhao, W., Wan, K., Li, Y.-J., Zhou, J., Miranda, D., Kale, A., \& Xu, C. (2024). Treat Visual Tokens as Text? But Your MLLM Only Needs Fewer Efforts to See. arXiv preprint arXiv:2410.05243.
\item
  Li, K. Y., Goyal, S., Dias Semedo, J., \& Kolter, J. Z. (2024). Inference Optimal VLMs Need Fewer Visual Tokens and More Parameters. arXiv preprint arXiv:2411.03312.
\item
  Wang, Y., Liu, C., Qu, Y., Cao, H., Jiang, D., \& Xu, L. (2024). Break the Visual Perception: Adversarial Attacks Targeting Encoded Visual Tokens of Large Vision-Language Models. arXiv preprint arXiv:2411.03312.
\end{enumerate}

\textbf{Figures}: 11 (1: PHI bbox to patch pipeline; 2: Coordinate
mapping; 3: DeepSeek-OCR architecture with hook points; 4: Compression
leakage schematic; 5: Masking variant convergence; 6: PHI masking
success by type; 7: Hybrid defense-in-depth flow; 8: Hybrid swimlane
PHI coverage; 9: Failure modes and mitigations; 10: Future work
roadmap; 11: Spatial coverage vs PHI reduction)

\textbf{Tables}: 2 (Table 1: Masking strategies; Table 2: Hybrid
pipeline)

\textbf{Code Availability}: Implementation (failed experiment) will be
archived on GitHub; URL to be added.

\textbf{Data Availability}: Synthetic data can be regenerated using
Synthea; scripts available in-repo.

\textbf{Competing Interests}: The authors declare no competing
interests.

\textbf{Funding}: None.

\end{document}